\documentclass[letterpaper, 10 pt, journal, twoside]{IEEEtran}      

\IEEEoverridecommandlockouts                              

\usepackage{amsmath,amssymb,amsfonts}

\usepackage{graphicx}
\usepackage{textcomp}
\usepackage{graphics} 
\usepackage{epsfig} 
\usepackage{subfigure}
\usepackage{amsmath} 
\usepackage{amssymb}  
\usepackage{epstopdf}
\usepackage{lipsum}
\usepackage{lipsum,amsmath,multicol}
\usepackage{float}
\usepackage{algorithm}
\usepackage{algpseudocode}
\usepackage{wrapfig}
\usepackage{sidecap}

\usepackage{amsthm}

\usepackage{comment}
\usepackage{array}
\usepackage{glossaries}
\usepackage{breqn}
\makeglossaries
\newcolumntype{L}[1]{>{\raggedright\let\newline\\\arraybackslash\hspace{0pt}}m{#1}}
\newcolumntype{C}[1]{>{\centering\let\newline\\\arraybackslash\hspace{0pt}}m{#1}}
\newcolumntype{R}[1]{>{\raggedleft\let\newline\\\arraybackslash\hspace{0pt}}m{#1}}
\usepackage{sidecap}
\usepackage{url}
\usepackage[T1]{fontenc}
\usepackage{tabularx}
\usepackage{multicol, blindtext}
\usepackage{amsmath}
\usepackage{multirow}
\usepackage{xcolor}
\usepackage{graphicx}
\usepackage{breqn}

\allowdisplaybreaks

\makeatletter
\def\endthebibliography{%
  \def\@noitemerr{\@latex@warning{Empty `thebibliography' environment}}%
  \endlist
}
\makeatother

\begin{document}
%
\title{Multi-Modal Model Predictive Control through Batch Non-Holonomic Trajectory Optimization: Application to Highway Driving}
\author{Vivek K. Adajania, Aditya Sharma, Anish Gupta, Houman Masnavi, K Madhava Krishna and Arun K.Singh 
\thanks{Manuscript received: September, 9, 2021; Revised December, 6, 2022; Accepted January, 13, 2022.}
\thanks{This paper was recommended for publication by Stephen J. Guy upon evaluation of the Associate Editor and Reviewers' comments.
This work was supported in part by the European Social Fund via IT Academy program in Estonia, smart specialization project with BOLT and grants COVSG24 and PSG605 from Estonian Research Council.}
\thanks{The first three authors are with the Robotics Research Center, Kohli Center for Intelligent Systems, IIIT Hyderabad, India and Houman Masnavi and Arun Singh are with the Institute of Technology, University of Tartu.}
\thanks{Digital Object Identifier (DOI): see top of this page.}
}

\markboth{IEEE Robotics and Automation Letters. Preprint Version. Accepted January, 2022}
{Adajania \MakeLowercase{\textit{et al.}}: Multi-Modal MODEL PREDICTIVE CONTROL} 
\maketitle



\begin{abstract}
Standard Model Predictive Control (MPC) or trajectory optimization approaches perform only a local search to solve a complex non-convex optimization problem. As a result, they cannot capture the multi-modal characteristic of human driving. A global optimizer can be a potential solution but is computationally intractable in a real-time setting. In this paper, we present a real-time MPC capable of searching over different driving modalities. Our basic idea is simple: we run several goal-directed parallel trajectory optimizations and score the resulting trajectories based on user-defined meta cost functions. This allows us to perform a search over several locally optimal motion plans. Although conceptually straightforward, realizing this idea in real-time with existing optimizers is highly challenging from technical and computational standpoints. With this motivation, we present a novel batch non-holonomic trajectory optimization whose underlying matrix algebra is easily parallelizable across problem instances and reduces to computing large batch matrix-vector products. This structure, in turn, is achieved by deriving a linearization-free multi-convex reformulation of the non-holonomic kinematics and collision avoidance constraints.  We extensively validate our approach using both synthetic and real data sets (NGSIM) of traffic scenarios. We highlight how our algorithm automatically takes lane-change and overtaking decisions based on the defined meta cost function. Our batch optimizer achieves trajectories with lower meta cost, up to 6x faster than competing baselines.


\end{abstract}

\vspace{-0.2cm}

\section{Introduction}


Human driving is a complex mixture of discrete level decisions (merge, overtake, etc.) and lower-level motion commands \cite{comb_ad_planning}.  If we adopt an optimization perspective, the multiple discrete decisions can be seen as local minima associated with the underlying non-convex trajectory optimization problem \cite{comb_ad_planning} \cite{ad_logical_const}. Local optimizers based on Sequential Quadratic Programming (SQP) or Gradient Descent (GD) are not equipped to search over all the local minima. On the other hand, global optimization techniques like mixed-integer programming  \cite{ad_logical_const} offer a potential solution but are not particularly useful in a real-time setting, especially in dense traffic scenarios. 


\begin{figure}[t]
  \includegraphics[scale=0.36]{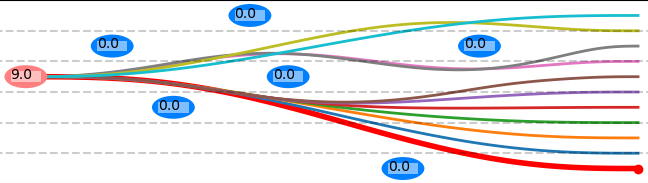}
  \caption{The different colored samples represent the several locally optimal trajectories for driving as close as possible to the cruise speed (meta cost). The ego-vehicle is shown in red while the neighboring obstacles are shown in blue. The number within the ellipses represent the velocities of the respective entity. Unlike \cite{fernet_planner}, \cite{behavior_ad}, we explicitly consider collision avoidance and kinematic constraints while generating candidate trajectories. The trajectory shown in bold red achieves the best performance (lowest meta-cost). We recommend seeing the accompanying video (\protect\url{https://tinyurl.com/3wew7vu7}) before reading the paper. }
  \label{hsrl_comb}
  \vspace{-0.8cm}
\end{figure}




\noindent \textbf{Main Idea:} Let us define goals as tuples of position, velocity and acceleration to be achieved by the ego-vehicle at the end of the planning horizon. Then, our approach in this paper is built on a simple insight that many different goal-directed  trajectories can accomplish a given high-level driving task. For example, if the task is to drive close to maximum velocity, the autonomous car can choose its next goal to be in any lane. Some particular goal choices may require overtaking a slow-moving vehicle directly in front, while some may require safely merging with oncoming cars in a different lane. More concretely, each goal-directed trajectories may converge to a different local minima resulting in a multi-modal driving behavior (see pp-5 \cite{comb_ad_planning}). Thus our proposed work is based on the idea of running several parallel goal-directed trajectory optimization problems and ranking the resulting locally optimal trajectories based on a user-defined meta cost function. Although conceptually simple, we are not aware of any such approach in existing works. There are trajectory sampling approaches such as \cite{fernet_planner}, \cite{behavior_ad} but they ignore collision avoidance and kinematic constraints while computing possible maneuvers. One possible reason the parallel/batch trajectory optimization approach has not been tried is that solving non-convex problem associated with autonomous driving is challenging. Running several instances of the problem in parallel only further increases the computational burden. The conceptually straightforward approach of running different optimizations in parallel CPU threads is not scalable for a large batch size in a dense highway driving scenario (see Fig.\ref{comp_time_batch} and discussions around it). We present a possible solution whose main novelties can be summarized as follows.


\noindent \textbf{Algorithmic:} We present the very first batch non-holonomic trajectory optimizer for real-time generation of several goal-directed locally optimal collision-free trajectories in parallel. The core algorithmic challenge lies in achieving linear scalability with respect to the batch size. As shown in Fig.\ref{comp_time_batch}(b),  we cannot achieve such scalability by simply running each optimization problem in a separate CPU thread. Instead, we need to parallelize the batch optimizer's per-iteration computation efficiently. We address the scalability issue by developing a batch optimizer wherein iterating over the different problem instances in parallel boils down to multiplying a single constant matrix with a set of vectors. We show that the heavily vectorized structure of our optimizer stems naturally from two key algorithmic developments. First, we adopt a linearization-free multi-convex reformulation of the kinematic and collision avoidance constraints. Second, we apply the Alternating Minimization (AM) technique to solve the reformulated problem. We rank the output of batch optimization with some simple meta cost functions that model the higher-level driving objectives (e.g., driving with high-speed). We show that the ranking mechanism coupled with an intelligent goal-sampling approach automatically leads to discovering lane-change, vehicle following, overtaking maneuvers based on the traffic scenario. Refer Section \ref{rel_work} for a summary of algorithmic contribution over the author's prior work.

\noindent \textbf{Applied:} We provide an open source implementation \cite{github_repo_driving} for review and to promote further research in this direction. 




\noindent \textbf{State-of-the-art Performance:}  We compare our batch optimizer based MPC with three strong baselines and show that we achieve better solutions (in terms of meta-cost value) while being up to 6x faster. Our first baseline is a standard MPC that computes just a single locally optimal trajectory. Our second baseline is batch multi-threaded implementation of optimal control solver ACADO \cite{acado}. Our final baseline is based on trajectory sampling in the Frenet frame \cite{fernet_planner}.



\begin{table}[!t]
\scriptsize
\centering
\caption{Important Symbols  } \label{symbols}
\begin{tabular}{|p{3.2cm}|p{4cm}|p{5cm}|p{5cm}|}\hline
$x_l(t), y_l(t), \psi_l(t)$ & Position and heading of the ego-vehicle at time $t$. \\\hline
$\xi_{xj}(t), \xi_{yj}(t)$ & Position of the $j^{th}$ obstacle at time $t$\\ \hline
\mbox{$\alpha_{j, l}(t), d_{j, l}(t)$ }& Variables associated with our collision avoidance model. Refer to text for details.\\ \hline
\end{tabular}
\vspace{-0.6cm}
\normalsize
\end{table}

\section{Preliminaries and Related Work}

\subsection{Symbols and Notations}
\noindent Small-case normal and bold font letters will be used to denote scalars and vectors respectively. Bold-font upper-case letters will represent matrices. The superscript $T$ will denote transpose of a matrix and vector. 
Some of the main symbols are summarized in Table \ref{symbols} while some are also defined in their first place of use. 


\subsection{Batch Non-Holonomic Trajectory Optimization} \label{rel_work}
\noindent We are interested in solving $l$ non-holonomic trajectory optimizations in parallel each of which can be formulated in the following manner. The resulting trajectories from the parallel problems will be ranked based on a meta cost function discussed later.

\vspace{-0.3cm}
\small
\begin{subequations}
\begin{align}
    \min \sum_t{\ddot{x}_l(t)^2+\ddot{y}_l(t)^2+\ddot{\psi}_l(t)^2}\label{cost_standard} \\
    \dot{x}_l(t) = v_l(t)\cos\psi_l(t), \dot{y}_l(t)=v_l(t)\sin\psi_l(t), \forall t  \label{nonhol_standard} \\
    (x_l(t), y_l(t), \psi_l(t))\in \mathcal{C}_{b,l} \label{boundary}\\
    v_{min} \leq v_l(t) \leq v_{max}, \sqrt{\ddot{x}_l(t)^2+\ddot{y}_l(t)^2}\leq a_{max}  \label{v_bound}\\
    -\frac{(x_l(t)-\xi_{xj}(t))^2}{a^2}-\frac{(y_l(t)-\xi_{yj}(t))^2}{b^2}+1\leq 0, \label{coll_nonhol_standar}
\end{align}
\end{subequations}
\normalsize


\noindent The subscript $l$ denotes that the specific variable belongs to the $l^{th}$ instance of the problem in the batch. The variables of the trajectory optimization are $(x_l(t), y_l(t), \psi_l(t), v_l(t))$. The cost function minimizes the squared acceleration value for the linear and angular motions. The equality constraints (\ref{nonhol_standard}) stems from the non-holonomic kinematics of the car. Constraints (\ref{boundary}) ensures the boundary conditions on the position, heading angle and their derivatives. The inequalities (\ref{v_bound}) represent the bounds on the forward velocities and total acceleration. The set of constraints (\ref{coll_nonhol_standar}) enforces the collision avoidance between the ego and the neighboring vehicles with the assumption that both are represented by road center-line aligned ellipses. For the ease of exposition and without loss of generality, we assume that every obstacle ellipse has the same major ($a$) and minor axis ($b$) dimension. The ($a, b$) includes the inflation to account for the size of the ego-vehicle. The ellipse of the ego-vehicle and obstacle will not overlap as long the minimum separation distance is greater than $\sqrt{a^2+b^2}$. It is worth pointing out that $a, b$ will be larger than the length and width of the cars. In other words, there will be some over-approximation of the ego-vehicle and obstacle footprints. To keep this value limited, we enforce some restriction on the heading of the ego-vehicle. We discuss this more in the beginning of Section III.

The typical control inputs for the ego-vehicle are the acceleration and the steering inputs. The former can be obtained by the derivative or finite difference of $v(t)$. The steering angle is given by $\arctan(\frac{\dot{\psi}(t)*h}{v(t)})$, where $h$ is the inter-axle distance \cite{alonso_mora_parallel}. 


\noindent \textbf{Note:} All $l$ instances of (\ref{cost_standard})-(\ref{coll_nonhol_standar}) have the same velocity and acceleration bounds and neighboring obstacles.

\noindent\textbf{Existing Works:} Trajectory optimizations of the form (\ref{cost_standard})-(\ref{coll_nonhol_standar}) are typically solved through approaches like SQP \cite{alonso_mora_parallel}, GD \cite{gd_ad} etc. Authors in \cite{ad_logical_const} proposed a global optimization approach but considered the ego vehicle as a holonomic triple-integrator system. They derived some approximations for the non-holonomic constraints that hold more naturally at high forward velocities. In contrast, \cite{knoll_ad} adopts a more rigorous approach but the resulting algorithm was tested on environments sparsely filled with obstacles. Authors in \cite{2stage_opt} used an approach similar to \cite{ad_logical_const} to compute the best driving modality and then refined the solution further through local optimization (e.g SQP) to handle kinematic constraints.


\noindent \textbf{Improvements over our prior efforts:} Our approach extends recent work \cite{aks_iros21} to batch setting and further applies it to highway driving. Specifically, we propose the core batch solution update rule (Eqn.(\ref{lin_solve_xy_2}), (\ref{lin_solve_psi_2}) ) by leveraging the implicit structures in the matrix algebra of \cite{aks_iros21}. Furthermore, unlike our current work, \cite{aks_iros21} did not consider acceleration bounds. Typically, constraints on acceleration are modeled as affine inequalities. However, such representation is not suitable for our formulation that relies on reducing the trajectory optimization to a sequence of unconstrained QPs to obtain an efficient batch update rule. We thus reformulate the acceleration bounds in the same form as the collision avoidance constraints of \cite{aks_iros21}. Our current work is also an improvement over \cite{aks_icra20} that handled collision avoidance constraints through a novel linearization approach. But as discussed in Section \ref{parallel_least_setting}, any linearization of the underlying costs and constraints substantially complicates the parallelization of the batch optimization.

\section{Main Results}
This section derives our main algorithmic results. We begin by summarizing our main assumptions.


\noindent \textbf{1.Road Attached Frame:} We assume that our problem set-up (i.e optimization (\ref{cost_standard})-(\ref{coll_nonhol_standar})) is defined in the reference frame of the center-line of the road \cite{2stage_opt}. This allows us to essentially treat curved roads as ones with a straight-line geometry. The non-holonomic constraints are defined in the road attached frame and holds true when the trajectories are reverted back to global frame \cite{2stage_opt}.

\noindent \textbf{2. Restricted Heading Change:} We post-process and discard locally optimal trajectories resulting from (\ref{cost_standard})-(\ref{coll_nonhol_standar}) that incur a large heading change ($\approx\pm13 deg.$) with respect to the road center-line. The said heading restriction is realistic in the case of typical highway driving scenarios \cite{nikolce_ad} and is made to accommodate the collision avoidance model defined in (\ref{coll_nonhol_standar}).

\noindent \textbf{3. Trajectory Prediction:} The batch optimization generates trajectories that are feasible with respect to the constant-velocity prediction of the trajectories of the  dynamic obstacles (neighboring cars). That is, $\xi_{x, j}(t), \xi_{y, j}(t)$ are obtained by linearly interpolating the positions with the current velocity. We adopt such a minimalist representation to test the full potential of our batch optimization, specifically how its fast re-planning ensures safety in the absence of any complex trajectory forecasting algorithms.

\subsection{Intuition from Parallel Least Squares}\label{parallel_least_setting}
\noindent Consider the following $l$ linear least squares problem

\vspace{-0.2cm}
\small
\begin{align}
    \min_{\textbf{s}_l}\Vert \textbf{F}\textbf{s}_l-\textbf{g}_l\Vert_2, \forall l = 1, 2....l
    \label{ls}
\end{align}
\normalsize

\noindent All problem instances share the same matrix $\textbf{F}$ but have different vector $\textbf{g}_l$. A simple way to solve the problem would be first to compute the inverse of $\textbf{F}^T\textbf{F}$, then calculate $\textbf{F}^T\textbf{g}_l$ and finally multiply both the entities together. Notably, the last two operations are matrix-vector products that can be trivially parallelized. The expensive inverse needs to be done only \textbf{once}. 

Now, contrast the above with the following non-linear least squares set-up

\vspace{-0.2cm}
\small
\begin{align}
    \min_{\textbf{s}_l}\Vert \textbf{f}(\textbf{s}_l)\Vert_2^2 \approx \min_{\textbf{s}_l}\Vert \textbf{F}_l\textbf{s}_l-\textbf{g}_l\Vert_2
    \label{nls}
\end{align}
\normalsize

\noindent The solution process begins by computing the Jacobian of $\textbf{f}$ around a given guess solutions $\textbf{s}_l$ to obtain a linear least squares approximation. Importantly, since each $\textbf{s}_l$ would be different, the matrix $\textbf{F}_l$ will vary across problem instances. Thus, computing the solution would first require forming $\textbf{F}_l^T\textbf{F}_l$ followed by computing inverses (or just factorization) of each of these. Furthermore, these computations need to be refined at each iteration of the non-linear least squares. 


The above example illustrates the relative difficulty of parallelizing the per-iteration computations of a non-linear least squares problem compared to the special linear setting presented in (\ref{ls}). Many trajectory optimizations are indeed formulated as non-linear least-squares \cite{teb_planner}, and thus, they inherit the same bottlenecks discussed above towards parallelization. We note that in the non-linear setting, instead of parallelizing per-iteration operations, it will be more reasonable to solve each non-linear least squares in full in parallel CPU threads. 



The core feature of our batch optimizer is that its most computationally heavy part has the same structure as the linear least-squares set-up of (\ref{ls}). Thus our parallelization effort essentially reduces to computing batch matrix-vector products.


\subsection{Building Blocks}

\noindent \textbf{Reformulating the Collision Avoidance Constraints:} We rephrase the quadratic collision avoidance constraints (\ref{coll_nonhol_standar}) into the form $\textbf{f}_{c, l} = \textbf{0}$ based on our prior works \cite{aks_iros21}.

\small
\begin{align}
    \textbf{f}_{c, l} = \left \{ \begin{array}{lcr}
x_l(t) -\xi_{xj}(t)-ad_{j,l}(t)\cos\alpha_{j,l}(t), \forall j, t \\
y_l(t) -\xi_{yj}(t)-bd_{j,l}(t)\sin\alpha_{j,l}(t), \forall j, t \\ 
\end{array} \right \}
\label{sphere_proposed}
\end{align}
\normalsize
\normalsize

\noindent As evident, $\textbf{f}_{c, l}$ resembles a polar representation of the Euclidean distance between the robot and the obstacle with the variables $\alpha_{j,l}(t)$ being the angle of the line of sight vector connecting the ego-vehicle and the $j^{th}$ obstacle. The variable $d_{j,l}(t)$ is the ratio of the length of the line-of-sight vector to minimum separation distance required for collision avoidance. Note that these variables will be different for each problem in the batch and the additional subscript $l$ has been introduced to represent that fact. Collision avoidance can be enforced by ensuring $d_{j, l}(t)\geq 1$. It should be noted that in (\ref{sphere_proposed}), $\alpha_{j,l}(t), d_{j, l}(t)$ are unknown variables that are obtained by our optimizer along with other trajectory variables.

\noindent \textbf{Reformulating Acceleration Bounds:} Typically, the quadratic acceleration bounds in (\ref{v_bound}) are split into separate affine constraints along each motion direction. However, we reformulate it in the same manner as collision avoidance constraints. That is we have constraints of the form $\textbf{f}_{a, l}=\textbf{0}$, where
\small
\begin{align}
    \textbf{f}_{a, l} = \left \{ \begin{array}{lcr}
\ddot{x}_l(t) -d_{a, l}(t)\cos\alpha_{a, l}(t) \\
\ddot{y}_l(t) -d_{a, l}(t)\sin\alpha_{a, l}(t)\\ 
\end{array} \right \}, , d_{a, l}(t)\leq a_{max}
\label{acc_bound_proposed}
\end{align}
\normalsize

\noindent The variables $\alpha_{a, l}(t), d_{a, l} $ are unknown and will be computed by our optimizer. On the surface, our representation of acceleration bounds seem more complicated but as shown later, is essential achieve appropriate computational structure in our batch optimizer.

\noindent \textbf{Trajectory Parameterization}
\noindent The trajectory variables in each of the $l$ instantiations of the problem  can be represented in the following manner.

\small
\begin{align}
    \begin{bmatrix}
        x(t_1), \dots, x(t_n) 
    \end{bmatrix} = \textbf{P}\textbf{c}_{x, l},
     \begin{bmatrix}
        \psi(t_1), \dots, \psi(t_n) 
    \end{bmatrix} = \textbf{P}\textbf{c}_{\psi, l}
    \label{param}
\end{align}
\normalsize

\noindent where, $\textbf{P}$ is a matrix formed with time-dependent basis functions (e.g polynomials) and $\textbf{c}_{x, l}, \textbf{c}_{\psi, l}$ are the coefficients associated with the basis functions. Similar expressions can be written for $y_l(t)$ as well. We can also express the derivatives in terms of $\dot{\textbf{P}}, \ddot{\textbf{P}}$. Our representation creates a low dimensional parametrization of the trajectories and has its own pros and cons. On one hand, it reduces the size of the optimization problem but on the other it also leads to less free parameters for the satisfaction of the constraints. We have observed that the representation (\ref{param}) has performed well with our multi-convex approximation-based approach for trajectory optimization for not only the current work but also prior efforts like \cite{aks_icra20}. However, it is unclear how well our low dimensional representation will work with standard SQP or GD-based approaches.

\noindent \textbf{Matrix Representation:} Using the trajectory parametrization presented in (\ref{param}), we can put constraints (\ref{sphere_proposed}) and (\ref{acc_bound_proposed}) and the non-holonomic constraints (\ref{nonhol_standard}) in the matrix form of (\ref{obs_matrix}), (\ref{acc_matrix}), (\ref{nonhol_matrix}) respectively.

\small
\begin{subequations}
\begin{align}
    \textbf{F}_o \textbf{c}_{x, l} = \boldsymbol{\xi}_x+a \textbf{d}_{l}\cos\boldsymbol{\alpha}_l,
    \textbf{F}_o \textbf{c}_{y, l} = \boldsymbol{\xi}_y+b \textbf{d}_l\sin\boldsymbol{\alpha}_l
    \label{obs_matrix}\\
    \ddot{\textbf{P}}\textbf{c}_{x, l} = \textbf{d}_{a, l}\cos\boldsymbol{\alpha}_a,
    \ddot{\textbf{P}} \textbf{c}_{y, l} = \textbf{d}_{a, l}\sin\boldsymbol{\alpha}_a, 
    \label{acc_matrix}\\
\dot{\textbf{P}}\textbf{c}_{x, l} = \textbf{v}_l\cos\boldsymbol{\textbf{P}\textbf{c}_{\psi, l}},
    \dot{\textbf{P}} \textbf{c}_{y, l} = \textbf{v}_l\sin\boldsymbol{\textbf{P}\textbf{c}_{\psi, l}}
    \label{nonhol_matrix}
\end{align}
\end{subequations}
\normalsize

\noindent The matrix $\textbf{F}_o$ is obtained by stacking the matrix $\textbf{P}$ from (\ref{param}) $m$ times (the number of obstacles in the 
environment). The vector $\boldsymbol{\xi}_{x}, \boldsymbol{\xi}_{y}$ is formed by appropriately stacking $\xi_{xj}(t), \xi_{yj}(t)$ at different time instants and for all the obstacles. Similar construction is followed to obtain $\boldsymbol{\alpha}_{l},  \boldsymbol{\alpha}_{a, l}, \textbf{d}_{l}, \boldsymbol{d}_{a, l}, \boldsymbol{\psi}_{l}, \textbf{v}_l$.


\subsection{Multi-Convex Reformulation}
\noindent Using previous derivations, we substitute (\ref{cost_standard})-(\ref{coll_nonhol_standar}) by the following:



\small
\begin{subequations}
\begin{align}
    \min_{v_{min} \leq \textbf{v}_l \leq \textbf{v}_{max}} \frac{1}{2}\textbf{c}_{x, l}^T\textbf{Q}\textbf{c}_{x, l}+\frac{1}{2}\textbf{c}_{y, l}^T\textbf{Q}\textbf{c}_{y, l}+\frac{1}{2}\textbf{c}_{\psi, l}^T\textbf{Q}\textbf{c}_{\psi, l} \label{cost_reform}  \\
    \textbf{A}\begin{bmatrix}
        \textbf{c}_{x, l}\\
        \textbf{c}_{y, l}
    \end{bmatrix}= \textbf{b}_l, \textbf{A} \textbf{c}_{\psi, l} = \textbf{b}_{\psi, l} \label{eq_reform} \\
    \textbf{F} \begin{bmatrix}
        \textbf{c}_{x, l}\\
        \textbf{c}_{y, l}\\
    \end{bmatrix} = \textbf{g}_l(\textbf{c}_{\psi, l}, \boldsymbol{\alpha}_{l}, \boldsymbol{\alpha}_{a,l}, \textbf{d}_{l}, \textbf{d}_{a, l}, \textbf{v}_l) \label{nonconvex_reform}  
\end{align}
\end{subequations}
\normalsize

\small
\begin{align}
    \textbf{F} = \begin{bmatrix}
    \begin{bmatrix}
    \textbf{F}_{o}\\
    \ddot{\textbf{P}}\\
    \dot{\textbf{P}}
    \end{bmatrix} & \textbf{0}\\
    \textbf{0} & \begin{bmatrix}
    \textbf{F}_{o}\\
    \ddot{\textbf{P}}\\
    \dot{\textbf{P}}
    \end{bmatrix} 
    \end{bmatrix}, \textbf{g}_l = \begin{bmatrix}
    \boldsymbol{\xi}_x+a \textbf{d}_{l}\cos\boldsymbol{\alpha}_{l}\\
     \textbf{d}_{a, l}\cos\boldsymbol{\alpha}_{a, l}\\
   \textbf{v}_l\cos\textbf{P}\textbf{c}_{\psi, l}\\
   \boldsymbol{\xi}_y+b \textbf{d}_{l}\sin\boldsymbol{\alpha}_{l}\\
     \textbf{d}_{a, l}\sin\boldsymbol{\alpha}_{a, l}\\
   \textbf{v}_l\sin\textbf{P}\textbf{c}_{\psi, l}\\
    \end{bmatrix},
\end{align}
\normalsize


\noindent The cost function is a matrix representation of the sum of squared acceleration term in (\ref{cost_standard}). The equality constraints (\ref{eq_reform}) are the matrix representation of the boundary constraints (\ref{boundary}). We stack all the non-convex equality constraints in (\ref{nonconvex_reform}). 

\newtheorem{remark}{Remark}
\begin{remark} \label{const_matrix_rep}
The matrices $(\textbf{Q}, \textbf{F}, \textbf{F}_o, \textbf{P}, \dot{\textbf{P}}, \ddot{\textbf{P}})$ in the reformulated problem (\ref{cost_reform})-(\ref{nonconvex_reform}) do not depend on the batch index $l$. In other words, they are the same for all the problem instantiations.
\end{remark}

\noindent Remark \ref{const_matrix_rep} highlights the motivation behind choosing the specific representation of the collision avoidance (\ref{sphere_proposed}) and acceleration bounds (\ref{acc_bound_proposed}).




\subsection{Solution by Alternating Minimization}
\noindent We solve (\ref{cost_reform})-(\ref{nonconvex_reform}) by relaxing the non-convex equality constraints (\ref{nonconvex_reform}) as $l_2$ penalties and augmenting them into the cost function.
\begin{dmath}
    f_{xy}(\textbf{c}_{x,l}, \textbf{c}_{y, l}, \boldsymbol{\lambda}_{x, l}, \boldsymbol{\lambda}_{y, l} )+f_{\psi}(\textbf{c}_{\psi, l}, \boldsymbol{\lambda}_{\psi,l} )+\frac{\rho_{xy}}{2}\left\Vert \textbf{F} \begin{bmatrix}
        \textbf{c}_{x, l}\\
        \textbf{c}_{y, l}
    \end{bmatrix} -\textbf{g}_l \right\Vert_2^2
    \label{aug_cost}
\end{dmath}
\normalsize
\vspace{-0.5cm}
\small
\begin{subequations}
\begin{align}
    f_{xy} = \frac{1}{2}\textbf{c}_{x, l}^T\textbf{Q}\textbf{c}_{x, l}+\frac{1}{2}\textbf{c}_{y, l}^T\textbf{Q}\textbf{c}_{y, l}-\langle \boldsymbol{\lambda}_{x, l}, \textbf{c}_{x, l}\rangle-\langle \boldsymbol{\lambda}_{y, l}, \textbf{c}_{y, l}\rangle\\
    f_{\psi} = \frac{1}{2}\textbf{c}_{\psi, l}^T\textbf{Q}\textbf{c}_{\psi, l}-\langle \boldsymbol{\lambda}_{\psi, l}, \textbf{c}_{\psi, l}\rangle
\end{align}
\end{subequations}
\normalsize

\begin{algorithm*}[!h]
\centering
 \caption{Alternating Minimization based Batch Non-Holonomic Trajectory Optimization }\label{algo_1}
    \begin{algorithmic}[1]   
\small
\While{$k \leq maxiter$}
\begin{align}
    {^{k+1}}(\textbf{c}_{x, l}, \textbf{c}_{y, l}) = \arg\min_{\textbf{c}_{x, l}, \textbf{c}_{y, l}} 
         f(\textbf{c}_x, \textbf{c}_y, {^k}\boldsymbol{\lambda}_{x, l}, {^k}\boldsymbol{\lambda}_{y, l} )+\frac{\rho_{xy}}{2}\left\Vert \textbf{F}\begin{bmatrix} \textbf{c}_{x, l}\\
    \textbf{c}_{y, l}
    \end{bmatrix} -\textbf{g} ({^{k}}\boldsymbol{\alpha}_{l}, {^k}\boldsymbol{\alpha}_{a, l}, {^k}\textbf{c}_{\psi, l}, {^k}\textbf{d}_{l}, {^k}\textbf{d}_{a, l}, {^k}\textbf{v}_l) \right \Vert_2^2, \textbf{A} \begin{bmatrix}
            \textbf{c}_{x, l}\\
            \textbf{c}_{y, l}
        \end{bmatrix} = \textbf{b}_l 
    \label{split_xy}
\end{align}
\normalsize

    
\small
\begin{align}
    {^{k+1}}\textbf{c}_{\psi} = \arg\min_{\textbf{A}\textbf{c}_{\psi, l} = \textbf{b}_{\psi, l}} f(\textbf{c}_{\psi})+\frac{\rho_{xy}}{2}\Vert \textbf{F} \begin{bmatrix}
        {^{k+1}}\textbf{c}_{x, l}\\
        {^{k+1}}\textbf{c}_{y,l}
    \end{bmatrix} -\textbf{g}(\textbf{c}_{\psi, l}) \Vert_2^2
    = \arg\min_{\textbf{A}\textbf{c}_{\psi,l}=\textbf{b}_{\psi, l}} f(\textbf{c}_{\psi}) + \frac{\rho_{xy}}{2}\left\Vert \begin{matrix}
    {^{k+1}}\dot{\textbf{x}}_l-{^k}\textbf{v}_l\cos\textbf{P}\textbf{c}_{\psi, l}\\
    {^{k+1}}\dot{\textbf{y}}_l-{^k}\textbf{v}_l\sin\textbf{P}\textbf{c}_{\psi, l}
    \end{matrix}\right \Vert_2^2 \nonumber \\ 
    = \arg\min_{\textbf{A}\textbf{c}_{\psi,l}=\textbf{b}_{\psi, l}} f(\textbf{c}_{\psi})+ \frac{\rho_{xy}}{2}\Vert \arctan2({^{k+1}}\dot{\textbf{y}}_l, {^{k+1}}\dot{\textbf{x}}_l)-\textbf{P}\textbf{c}_{\psi, l} \Vert_2^2
    \label{split_psi}
\end{align}
\normalsize


\small
\begin{align}
    {^{k+1}}\textbf{v}_l = \arg\min_{v_{min} \leq \textbf{v}_l \leq \textbf{v}_{max} } \left\Vert \textbf{F}\begin{bmatrix}
        {^{k+1}}\textbf{c}_{x, l}\\
        {^{k+1}}\textbf{c}_{y, l}\\
    \end{bmatrix}-\textbf{g}({^{k+1}}\textbf{c}_{\psi, l}, \textbf{v}_l)\right\Vert_2^2
    =\arg\min_{v_{min} \leq \textbf{v} \leq \textbf{v}_{max} } \left \Vert\begin{matrix}
    {^{k+1}}\dot{\textbf{x}}_l-\textbf{v}_l\cos\boldsymbol{\psi}_l\\
    {^{k+1}}\dot{\textbf{y}}_l-\textbf{v}_l\sin\boldsymbol{\psi}_l\\
    \end{matrix} \right\Vert_2^2 
    \label{split_v}
\end{align}
\normalsize

\small
\begin{align}
    {^{k+1}}\boldsymbol{\alpha}_l = \arg\min_{\boldsymbol{\alpha}_l}\left\Vert \textbf{F}\begin{bmatrix}
        {^{k+1}}\textbf{c}_{x, l}\\
        {^{k+1}}\textbf{c}_{y, l}
    \end{bmatrix}-\textbf{g}\right\Vert_2^2 = \arg\min_{\boldsymbol{\alpha}_l}\left \Vert \begin{matrix}
    {^{k+1}}\textbf{x}_l-\boldsymbol{\xi}_{x}-a\hspace{0.1cm}{^k}\textbf{d}_l\cos\boldsymbol{\alpha}_l\\
    {^{k+1}}\textbf{y}_l-\boldsymbol{\xi}_{y}-b\hspace{0.1cm}{^k}\textbf{d}_l\sin\boldsymbol{\alpha}_l\\
    \end{matrix}\right\Vert_2^2, {^{k+1}}\textbf{d}_l = \arg\min_{\textbf{d}_l\geq 1}\left\Vert \textbf{F}\begin{bmatrix}
        {^{k+1}}\textbf{c}_{x, l}\\
        {^{k+1}}\textbf{c}_{y, l}
    \end{bmatrix}-\textbf{g}({^{k+1}}\boldsymbol{\alpha}_l)\right\Vert_2^2
    \label{split_alpha_o}
\end{align}
\normalsize


\small
\begin{align}
    {^{k+1}}\boldsymbol{\alpha}_{a, l} = \arg\min_{\boldsymbol{\alpha}_{a, l}}\left\Vert \textbf{F}\begin{bmatrix}
        {^{k+1}}\textbf{c}_{x, l}\\
        {^{k+1}}\textbf{c}_{y, l}
    \end{bmatrix}-\textbf{g}({^k}\textbf{d}_{a, l}, \boldsymbol{\alpha}_{a, l})\right\Vert_2^2, {^{k+1}}\textbf{d}_{a, l} = \arg\min_{\textbf{d}_{a, l} \leq a_{max}} \left\Vert \textbf{F}\begin{bmatrix}
        {^{k+1}}\textbf{c}_{x, l}\\
        {^{k+1}}\textbf{c}_{y, l}
    \end{bmatrix}-\textbf{g}({^{k+1}}\boldsymbol{\alpha}_{a, l}, \textbf{d}_{a, l})\right\Vert_2^2
    \label{split_alpha_a}
\end{align}
\normalsize

\EndWhile
\end{algorithmic}
\end{algorithm*}

\noindent Note the introduction of so-called Lagrange multipliers $\boldsymbol{\lambda}_{x, l}$, $\boldsymbol{\lambda}_{y, l}$, and $\boldsymbol{\lambda}_{\psi, l}$ that play a crucial role in driving the residuals of the equality constraints to zero \cite{admm_neural}. 


\begin{remark}\label{multi_convex_block}
The augmented cost function (\ref{aug_cost}) is convex in $(\textbf{c}_{x,l}, \textbf{c}_{y, l})$ for a given $\textbf{c}_{\psi, l}, \boldsymbol{\alpha}_l, \boldsymbol{\alpha}_{a, l}, \textbf{d}_{l}, \textbf{d}_{a, l}$. Similarly, it is convex in $\textbf{d}_l, \textbf{d}_{a, l}$ for a given  $ \textbf{c}_{\psi, l}, \boldsymbol{\alpha}_l, \boldsymbol{\alpha}_{a, l}, \textbf{c}_{x, l}, \textbf{c}_{y, l}$
\end{remark}

\begin{remark} \label{conv_surrogate}
For a given $(\textbf{c}_{x, l}, \textbf{c}_{y, l})$, the cost function (\ref{aug_cost}) is non-convex in $\textbf{c}_{\psi, l} $ but can be replaced with a simple convex surrogate from \cite{aks_icra20}.
\end{remark}

\begin{remark} \label{closed_form}
For a given $(\textbf{c}_{x, l}, \textbf{c}_{y, l})$, the optimizations over variables $(\boldsymbol{\alpha}_l, \boldsymbol{\alpha}_{a, l})$ have a simple closed form solution.
\end{remark}

\begin{remark}\label{feasibility}
The augmented Lagrangian based reformulation of the non-convex constraints ensures that our batch optimization is always feasible. As a result, it can handle infeasible (e.g with respect to collision avoidance) trajectory initialization.
\end{remark}

\noindent Remarks \ref{multi_convex_block} and \ref{conv_surrogate} are precisely the multi-convex structure foreshadowed in the earlier sections. Moreover, remarks \ref{multi_convex_block}-\ref{closed_form} highlight why an AM approach would be suitable: by decomposing the optimization process over separate blocks of variable, we can leverage the implicit convex structures present in the problem. The use of Augmented Lagrangian cost (\ref{aug_cost}) in combination with AM procedure is known as the split-Bregman technique \cite{admm_neural}.

The different steps of AM are presented in (\ref{split_xy})-(\ref{split_alpha_a}), wherein the left superscript $k$ is used to track the value of the variable over different iterations. For example, ${^k}\textbf{c}_{x, l}$ represents the value at iteration $k$ of this specific variable. At each optimization block, only few specific variables are optimized while the rest are kept fixed at the values obtained in the previous iteration or the previous step of the same iteration.

\subsection{Analysis }

\noindent \textbf{Step (\ref{split_xy})} This optimization is a convex equality constrained QP that reduces to solving the following set of linear equations, wherein $\boldsymbol{\mu}_{x, l}, \boldsymbol{\mu}_{y, l}$ are the dual variable associated with the equality constraints.

\small
\begin{align}
\overbrace{\begin{bmatrix}
\textbf{Q}+\rho_{xy}\textbf{F}^T\textbf{F} & \textbf{A}^{T}\\
\textbf{A} & \textbf{0} 
\end{bmatrix}}^{\textbf{Q}_{xy}}
\begin{bmatrix}
    \textbf{c}_{x, l}\\
    \textbf{c}_{y, l}\\
    \boldsymbol{\mu}_{x, l}\\
    \boldsymbol{\mu}_{y, l}\\
    \end{bmatrix} = \overbrace{\begin{bmatrix}
    \rho_{xy}\textbf{F}^T {^k}\textbf{g}_l+\begin{bmatrix}
        {^k}\boldsymbol{\lambda}_{x, l}\\
        {^k}\boldsymbol{\lambda}_{y, l}\\
    \end{bmatrix}\\
    \textbf{b}_l
    \end{bmatrix}}^{\textbf{q}_l}
\label{lin_solve_xy_1}
 \end{align}
\normalsize

\noindent The set of equations (\ref{lin_solve_xy_1}) computes the solution for the $l^{th}$ instance of the problem. However, since the left hand side of (\ref{lin_solve_xy_1}) does not depend on the batch index $l$, we can compute the solution of the entire batch in one-shot through (\ref{lin_solve_xy_2}).

\small
\begin{dmath}
    \begin{bmatrix}
        \textbf{c}_{x, 1}, \textbf{c}_{y, 1}, \boldsymbol{\mu}_{x, 1}, \boldsymbol{\mu}_{y, 1}\\
        \vdots\\
        \textbf{c}_{x, l}, \textbf{c}_{y, l}, \boldsymbol{\mu}_{x, l}, \boldsymbol{\mu}_{y, l} 
    \end{bmatrix} = \left (\textbf{Q}_{xy}^{-1} \begin{bmatrix}
    \textbf{q}_1|  \textbf{q}_2 |\textbf{q}_1 \dots \textbf{q}_l
\end{bmatrix}\right )^T
\label{lin_solve_xy_2}
\end{dmath}
\normalsize

\noindent The major computation cost of (\ref{lin_solve_xy_2}) stems from obtaining different $\textbf{F}{^k}\textbf{g}_l$. But it is straightforward to formulate this operation as one large matrix-vector product and subsequently parallelize its computation. 

\noindent\textbf{Step (\ref{split_psi}) } As mentioned earlier, optimization over $(\textbf{c}_{\psi, l})$ is non-convex due to the presence of the non-holonomic penalty (second term). However, as shown in the last line of (\ref{split_psi}), for a given $({^{k+1}}\dot{\textbf{x}}_l {^{k+1}}\dot{\textbf{y}}_l)$, the non-convex term can be replaced with a convex surrogate over $\textbf{c}_{\psi, l}$, thereby reducing our problem to an equality-constrained QP. The solution process boils down to solving the following set of linear equations
\vspace{-0.5cm}
\small
\begin{dmath}
\overbrace{\begin{bmatrix}
\textbf{Q}+\rho_{xy}\textbf{P}^T\textbf{P} & \textbf{A}^{T}\\
\textbf{A} & \textbf{0} 
\end{bmatrix}}^{\textbf{Q}_{\psi}}
\begin{bmatrix}
    \textbf{c}_{\psi, l}\\
    \boldsymbol{\mu}_{\psi, l}
    \end{bmatrix} = \overbrace{\begin{bmatrix}
    \rho_{xy}\textbf{P}^T \arctan2({^{k+1}}\dot{\textbf{y}}_l, {^{k+1}}\dot{\textbf{x}}_l)+{^k}\boldsymbol{\lambda}_{\psi, l}\\
    \textbf{b}_{\psi, l}
    \end{bmatrix}}^{\textbf{q}_{\psi, l}}
\label{lin_solve_psi_1}
 \end{dmath}
\normalsize

\noindent Similar to the previous step, the left hand side of (\ref{lin_solve_psi_1}) do not depend on the batch index $l$ and thus we can compute the solution for the entire batch in one-shot

\small
\begin{dmath}
    \begin{bmatrix}
        \textbf{c}_{\psi, 1}, \boldsymbol{\mu}_{\psi, 1}\\
        \vdots\\
        \textbf{c}_{\psi, l}, \boldsymbol{\mu}_{\psi, l}\\
    \end{bmatrix} = \left (\textbf{Q}_{\psi}^{-1} \begin{bmatrix}
    \textbf{q}_{\psi, 1}|  \textbf{q}_{\psi, 2}|\textbf{q}_{\psi, 3} \dots \textbf{q}_{\psi, l}
\end{bmatrix}\right )^T
\label{lin_solve_psi_2}
\end{dmath}
\normalsize

\noindent\textbf{Step (\ref{split_v}):} For a given ${^{k+1}}\textbf{c}_{x, l}, {^{k+1}}\textbf{c}_{y, l}, {^{k+1}}\textbf{c}_{\psi, l}$ or alternately $({^{k+1}}\dot{\textbf{x}}_l, {^{k+1}}\dot{\textbf{y}}_l, {^{k+1}}\textbf{c}_{\psi, l} )$, the velocity $v_l(t)$ at different time instants can be treated as independent of each other. In other words, each element of $\textbf{v}_l$ is decoupled and thus the optimization (\ref{split_v}) reduces to $n$ parallel single-variable quadratic programming problems with a closed form solution. For $v_{min}>0$, the solution is given by (\ref{v_update}), wherein $clip(.)$ performs simple thresholding to satisfy the velocity bounds.
\small
\begin{align}
    {^{k+1}}\textbf{v}_l = clip(\sqrt{{^{k+1}}\dot{\textbf{x}}_l^2+{^{k+1}}\dot{\textbf{y}}_l^2}, v_{min}, v_{max})
    \label{v_update}
\end{align}
\normalsize

\noindent \textbf{First part of Step (\ref{split_alpha_o}):} For a given $({^{k+1}}{\textbf{x}}, {^{k+1}}{\textbf{y}})$, each element of $\boldsymbol{\alpha}_l$ can be considered to be decoupled from each other. Thus, the first optimization in (\ref{split_alpha_o}) separates into $n$ decoupled problems with the following closed form solution

\begin{align}
    {^{k+1}}\boldsymbol{\alpha}_l =  \arctan2(  a({^{k+1}} \textbf{y}_l-\boldsymbol{\xi}_y), b({^{k+1}} \textbf{x}_l-\boldsymbol{\xi}_x) )
\end{align}

\noindent \textbf{Second part of Step (\ref{split_alpha_o}):}: Similar to previous step, each element of $\textbf{d}_l$ can be considered independent and thus optimization over $\textbf{d}_l$ reduces to $n$ parallel single-variable QP with simple bound constraints. We can obtain a closed form solution by first solving the unconstrained problem and then simply clipping the value to lie between $[0 \hspace{0.1 cm} 1]$. 

\noindent \textbf{Step (\ref{split_alpha_a})}: The two optimizations in this step have the same structure as (\ref{split_alpha_o}) and thus have a closed form solution.

\begin{remark}\label{element-wise_op}
The solution of optimization (\ref{split_v})-(\ref{split_alpha_a}) involves only element-wise operations without any need of computing matrix factorization or inverse. Thus, computing a batch solution is trivial.
\end{remark}

\noindent\textbf{Multiplier Update:} The Lagrange multipliers are updated in the following manner \cite{admm_neural} which can be trivially done over the entire batch in one-shot.

\small
\begin{dmath}
    ({^{k+1}}\boldsymbol{\lambda}_{x, l}, {^{k+1}}\boldsymbol{\lambda}_{y, l}) = ({^{k}}\boldsymbol{\lambda}_{x, l}, {^{k}}\boldsymbol{\lambda}_{y, l})-\rho_{xy}\textbf{F}^T(\textbf{F}\begin{bmatrix}
        {^{k+1}}\textbf{c}_{x, l}\\
        {^{k+1}}\textbf{c}_{y, l}\\
    \end{bmatrix}-{^{k+1}}\textbf{g} )
\end{dmath}
\vspace{-0.4cm}
\small
\begin{align}
    {^{k+1}}\boldsymbol{\lambda}_{\psi, l} = {^{k}}\boldsymbol{\lambda}_{\psi, l}-\rho_{xy}\textbf{P}^T(\arctan2({^{k+1}}\dot{\textbf{y}}, {^{k+1}}\dot{\textbf{x}})-\textbf{P}{^{k+1}}\textbf{c}_{\psi, l})
\end{align}
\normalsize



\subsection{Goal Sampling and Meta-Cost}
\noindent This section provides a goal-sampling procedure for our batch optimizer and a meta-cost function to rank the resulting trajectories. We consider two typical scenarios encountered in highway driving.

\noindent \textbf{Cruise Driving:} Our first scenario considers driving forward with velocity as close as possible to a given $v_{cruise}$. Thus, our meta-cost is defined as simply.
\small
\begin{align}
    \sum_t (v(t)-v_{cruise})^2.
    \label{cruise_cost}
\end{align}
\normalsize
\noindent The goal position are spread evenly on different lanes, each at a distance of $v_{cruise}*t_f$, where $t_f$ is the planning horizon.



\noindent \textbf{Driving with Maximum Speed close to the Right Lane:} In this scenario, the ego-vehicle is required to drive as close as possible to maximum speed $v_{max}$ while being as close to the right-lane. The meta-cost is defined as the following wherein $y_{rl}$ is the lateral coordinate of the right-lane and $w_1$ and $w_2$ are user-defined constants.
\small
\begin{align}
    \sum_t w_1( v(t)-v_{max})^2+w_2(y(t)-y_{rl})^2.
    \label{hsrl_cost}
\end{align}
\normalsize 

\noindent The goals are sampled in the following manner. Around  $60 \%$ of the goals are placed on the right-lane at different distances. The remaining goals are spread across different lanes at a distance of $v_{max}*t_f$ from the current position. 


\section{Validation and Benchmarks}



    

\noindent \textbf{Implementation Details:}We implemented our batch optimizer in C++ using Eigen \cite{eigen}. We used $l = 11, a = 5.6, b = 3.1$ in the simulations. For each driving scenario discussed in the previous section, we created two variants depending on whether the neighboring vehicles follow the synthetic Intelligent Driver Model (IDM) or the pre-recorded trajectories from NGSIM data-set (6 different scenes) \cite{ngsim}. In the IDM data-set, each neighboring vehicle moves parallel to center-line and just adapt their cruise forward velocity based on the distance to the vehicles in front. In the NGSIM data set, the neighboring vehicles executes the pre-recorded trajectories. It is worth reiterating that our batch MPC and all the baselines have access to only the instantaneous position and velocity of the neighboring vehicles and not their true trajectories over the planning horizon. 

\noindent\textbf{Optimizer Convergence:} Fig.\ref{comp_time_batch}(a) shows the typical residual curve obtained with our batch optimizer in one of the MPC cycles. We show the trend for the best performing trajectory in the batch. We observed that on an average 100 iterations are enough to obtain residuals in the range of $10^{-3}$ for all the constraints.

\begin{table*}[h]
\centering
\caption{Meta-Cost Values in different Driving Scenarios (Mean/Min/Max). Lower is better}
\label{meta-cost_table} 
\scriptsize
\begin{tabular}{|l|l|l|l|l|}
\hline
Method & Cruise driving (IDM) & Cruise driving (NGSIM) & High-speed driving (IDM) & High-speed driving (NGSIM)   \\ \hline
Standard MPC & 5.41 / 0.0 / 50.97 & 3.260 / 0.0 / 54.3 & 2141.4 / 1360.8 / 2668.75 & 997.30 / 582.59 / 1256.1 \\ \hline
\textbf{Ours}  & \textbf{0.01 / 0.0 / 0.05} & \textbf{0.057 / 0.0 / 0.44} & \textbf{238.0 / 135.62 / 425.14} & \textbf{236.62 / 142.14 / 574.85} \\ \hline
ACADO batch size 11 & 0.08 / 0.0 / 0.66 & 0.114 / 0.0 / 0.62 & 381.96 / 183.01 / 1304.08 & 376.29 / 149.93 / 880.42 \\ \hline
ACADO batch size 6 & 0.12 / 0.0 / 1.06 & 0.103 / 0.0 / 0.85 & 643.97 / 129.5 / 1379.8 & 403.46 / 169.69 / 646.85 \\ \hline
frenet-frame planner & 0.14 / 0.0 / 1.00 & 0.280 / 0.02 / 0.95 & 563.71 / 194.64 / 1276.27 & 640.48 / 323.77 / 1144.0 \\ \hline
\end{tabular}
\normalsize
\end{table*}

\begin{table*}[h]
\centering
\caption{Acceleration Magnitudes Across Different Scenarios(Mean/Min/Max). Lower is better for Cruise driving. For high-speed driving, higher linear acceleration is better.}
\label{acc_table} 
\scriptsize
\begin{tabular}{|l|l|l|l|l|}
\hline
Method & Lin. Acc. Cruise driving & Lin. Acc. High-speed driving  & Ang. Acc. Cruise driving & Ang. Acc. High-speed driving   \\ \hline
Standard MPC & 0.93 / 0.00 / 2.63 & 0.72 / 0.00 / 1.56 & 0.02 / 0.00 / 0.07 & 0.01 / 0.00 / 0.03 \\ \hline
\textbf{Ours}  & \textbf{0.11 / 0.00 / 0.28} & \textbf{0.99 / 0.00 / 1.77} & \textbf{0.02 / 0.00 / 0.07} & \textbf{0.43 / 0.00 / 0.16}  \\ \hline
ACADO batch size 11 & 0.39 / 0.00 . 0.99 & 0.60 / 0.00 / 1.69 & 0.03 / 0.00 / 0.08 & 0.04 / 0.00 / 0.12 \\ \hline
ACADO batch size 6 & 0.43 / 0.00 / 1.10 & 0.57 / 0.00 / 1.69 & 0.03 / 0.00 / 0.07 & 0.03 / 0.00 / 0.12 \\ \hline
frenet-frame planner & 1.21 / 0.00 / 0.45 & 0.19 / 0.00 / 0.48 & 0.15 / 0.00 / 0.44 & 0.13 / 0.00 / 0.40 \\ \hline
\end{tabular}
\normalsize
\vspace{-0.4cm}
\end{table*}

\noindent \textbf{Baselines:} We benchmark against the following baselines
\begin{itemize}
    \item \textbf{Standard MPC}: We formulate a single batch MPC wherein the meta cost function is directly embedded into the trajectory optimizer to compute a single locally optimal trajectory. We use state-of-the-art optimal control framework ACADO \cite{acado} as the solver for the standard MPC. 
    
    \item \textbf{MPC with Batch ACADO:} We construct a batch version of ACADO, which solves several goal directed MPC over parallel CPU threads. This parallelization does not require any changes to be made in the matrix-algebra on the underlying SQP solver in ACADO. Thus, we use this set-up to highlight the computation gain resulting from our batch solver wherein the per-iteration computation itself vectorizes across problem instances. 
    
    \item \textbf{Frenet Frame Planner:} We also compare our batch optimizer with trajectory sampling approach presented in \cite{fernet_planner} which has been extensively used in the autonomous driving community and inspired similar related approaches like \cite{behavior_ad}. 
    

\end{itemize}

\subsection{Benchmarking}
\noindent \textbf{Cruise Scenario:} Table \ref{meta-cost_table} quantify the performance in the cruise-driving scenario obtained with different methods. The first two columns show the statistics of the velocity residuals $(v(t)-v_{cruise})^2$ observed over the full run of the MPC. The standard MPC performs the worst with a mean residual of $5.41$ and $3.26$ on the synthetic (IDM) and NGSIM data-set respectively. The worst-case performance is around $50$ on both data-sets. The performance is due to the fact in dense traffic scenarios, the standard MPC trajectories are unable to find a trajectory around the neighboring slow moving vehicles (see accompanying video). 

Our batch optimizer performs the best with a mean residual of $0.01$ and $0.05$ on the synthetic (IDM) and NGSIM data-set. In comparison, ACADO with a batch size of 11 and 6 shows comparable performance in terms of mean values. However, our batch optimizer achieves 12 times improvement over the worst-case numbers on the synthetic data set. The Frenet-frame planner with mean residuals of $0.14$ and $0.28$ performed worse than our's and batch ACADO. Its worst-case residual, though, is ten times ours.

\noindent\textbf{High Speed Driving with Right-Lane Preference:} This scenario has two competing terms in the meta cost: maximizing forward velocity and minimizing lateral distance to the right-lane. Thus, we adopt a slightly different analysis then before. We first compare the meta cost value across different methods and subsequently show how those translate to the physical metrics. Table \ref{meta-cost_table} (last two columns) summarizes the former results. As before, the standard MPC performs worst while our batch optimizer achieves the lowest meta cost value on both synthetic (IDM) and NGSIM data-set. We map these cost values to the achieved forward velocity and lateral distance residual in Table \ref{metric_hsrl}. For clarity, we present the combined results over the two data sets. We observe interesting  trends here as each approach attempts to minimize the meta-cost value by trading off velocities and lateral-distance residual in their own way. 


Our batch optimizer achieves highest mean forward velocity of $19.26 m/s$.  The performance of ACADO with a batch size of 11 is comparable to ours at $18.77 m/s$ while Frenet-frame planner's value stood substantially lower at $16.65 m/s$. 


Our batch optimizer also maintains a smaller lateral distance to the right-lane than parallelized ACADO. Our mean distance is $4.43m$ and in comparison, ACADO with batch size 11 managed a distance residual of $5.31 m$ on average. The respective values achieved with Frenet-frame planner is smaller than batch ACADO and even ours since it choose a smaller velocity to quickly converge to the right-lane.



\noindent \textbf{Acceleration Effort:} Table \ref{acc_table} presents the acceleration statistics observed across different driving scenarios. For ease of exposition, we combined the data obtained on the synthetic (IDM) and NGSIM data-set and present the overall mean, minimum and maximum. In the cruise driving scenario, the ideal acceleration is zero as the ego-vehicle is required to maintain a constant velocity. As can be seen from Table \ref{acc_table} (first column), our batch optimizer comes very close to the ideal performance with mean and maximum linear acceleration of $0.11 m/s^2$ and $0.28 m/s^2$ respectively. Both the mean and maximum value of the angular acceleration values are very close to zero. All the other approaches perform worse. Intuitively, a low acceleration value suggests that our batch optimizer could continuously navigate to free space less obstructed by neighboring vehicles. We highlight this explicitly in the accompanying video. All other approaches perform substantially worse. For example, the standard MPC's mean linear acceleration is over 9 times higher than ours. ACADO with batch size of 11 and 6 uses around 4 times higher acceleration magnitudes. Unsurprisingly, Frenet-frame planner performs worst since it ignores collision-avoidance constraints during trajectory generation process. Thus, it routinely enters a situation from where it needs to apply either emergency braking or execute a sharp turn to avoid collisions.

For high-speed driving, we want the ego vehicle to drive at max speed. Thus, in contrast to cruise-driving, here, large accelerations magnitude is indeed necessary for task fulfillment. As shown in Table \ref{metric_hsrl},  since our batch optimizer results in trajectories that gives more preference to maximizing speed, our mean accelerations are also highest. On the other hand, Frenet-frame planner achieved the lowest velocity and thus its linear acceleration values are also the lowest. 

\noindent \textbf{Computation Time:} We now present the most important result that is crucial in understanding the previous results in the appropriate context. Table \ref{comp-time-table} compares the computation-time of ours with all the baselines obtained on a $i7-8750$ processor with 16 GB RAM. Our batch optimizer with mean time of $0.07s$ is around $6\times$ faster than ACADO with a batch size of 11. In other words, ACADO needs substantially more computation budget to be even loosely competitive with our optimizer. Frenet-frame planner's timing is comparable to ours but as shown earlier, it performs the worst among all the multi-modal baselines in terms of meta-cost and acceleration effort. Our Frenet frame planner implementation used on an average 500 samples. Increasing this number, could improve the performance but at the expense of higher computation time. 

\begin{figure}[!h]
\centering
  \includegraphics[scale=0.11]{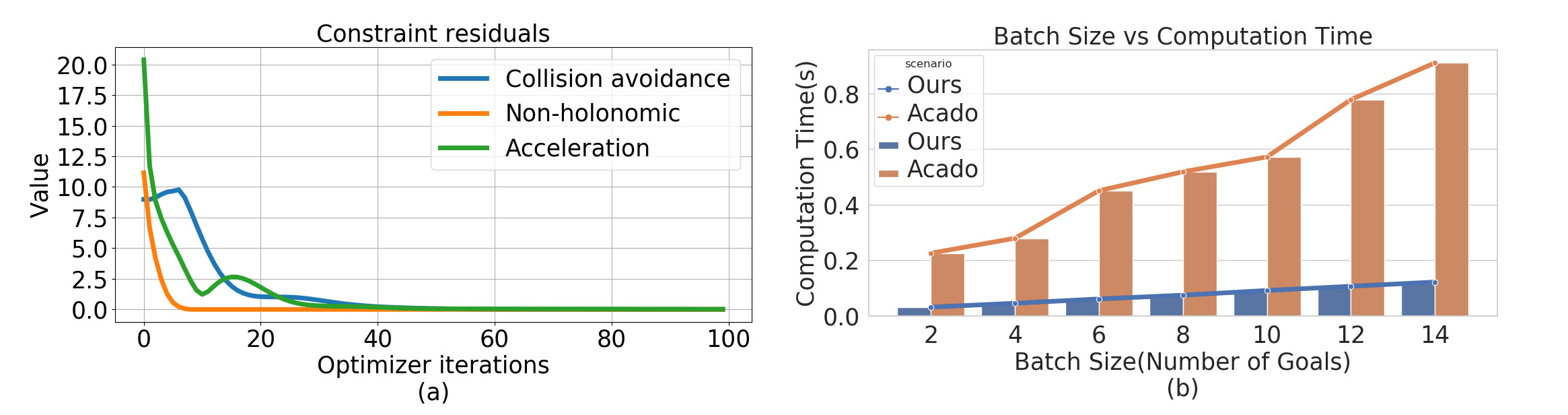}
  \caption{(a): Residual trend observed for best performing trajectory in a batch in one of the MPC cycles. We show the different components of constraint (\ref{nonconvex_reform}).  (b): Computation time vs Batch Size comparison between our batch optimizer and  batched ACADO \cite{acado} parallelized on multiple CPU threads. }
  \label{comp_time_batch}
\end{figure}

Fig.\ref{comp_time_batch} shows how computation time scales with batch size in our optimizer and batched multi-threaded ACADO. Our optimizer shows a linear increase which can be understood in the following manner. In eqn. (\ref{lin_solve_xy_2}), the matrix $\textbf{Q}_{xy}$ is independent of the batch size and only the matrix-vector on the r.h.s of (\ref{lin_solve_xy_2}) increases quadratically with it. The rate of increase can be made linear with simple parallelization of matrix multiplication. In contrast, batch ACADO solves full SQP in parallel CPU cores. Such parallelization efforts have thread synchronization overhead. Since the total cores in a laptop is typically 4 to 6, at batch size of 11 (or even 6), all SQP instantiations compete with each other for computing resources. Unfortunately, that is the best that we can achieve with off-the-shelf solvers. Since SQP relies on linearization, it isn't easy to parallelize its computation at each iteration (recall Section \ref{parallel_least_setting}).

\begin{table}[h]
\centering
\caption{Metrics for High-speed driving (Mean/Min/Max)}
\label{metric_hsrl}
\scriptsize
\begin{tabular}{|l|l|l|l|l|}
\hline
Method & Lat. dist. from right-lane  & Velocity\\ \hline
Standard MPC & 2.72 / 0.0 / 6.36 & 11.16 / 7.66 / 16.45 \\ \hline
\textbf{Ours}  & \textbf{4.43 / 0.0 / 18.0} & \textbf{19.28 / 17.14 / 20.68} \\ \hline
ACADO batch size 11 & 5.31 / 0.0 / 18.0 & 18.77 / 15.48 / 20.17  \\ \hline
ACADO batch size 6 & 3.5 / 0.0 / 15.43 & 17.64 / 14.10 / 20.56 \\ \hline
Frenet-frame planner & 2.42 / 0.0 / 9.34 & 16.65 / 14.0 / 19.74 \\ \hline
\end{tabular}
\normalsize
\vspace{-0.4cm}
\end{table}

\begin{table}[h]
\centering
\caption{MPC computation time[s] (Mean/Min/Max)}
\scriptsize
\label{comp-time-table} 
\begin{tabular}{|l|l|l|l|l|}
\hline
Method & Cruise driving  & High-speed driving\\ \hline
Standard MPC & 0.36 / 0.05 / 0.62 & 0.28 / 0.05 / 0.61 \\ \hline
\textbf{Ours}  & \textbf{0.07 / 0.06 / 0.08} & \textbf{0.07 / 0.06 / 0.07} \\ \hline
ACADO batch size 11 & 0.36 / 0.10 / 0.61 & 0.44 / 0.20 / 0.67  \\ \hline
ACADO batch size 6 (infeasible) & 0.25 / 0.11 / 0.39 & 0.32 / 0.15 / 0.47 \\ \hline
Frenet-frame planner & 0.082 / 0.068 / 0.097 & 0.08 / 0.07 / 0.096 \\ \hline
\end{tabular}
\vspace{-0.4cm}
\end{table}
\normalsize

\section{Discussions and Future Work}

This paper showed how a standard MPC based on local optimization techniques cannot generate sophisticated driving behaviors. Past works such as \cite{fernet_planner} have attempted to capture the multi-modality of autonomous driving by generating many candidate trajectories to different goals and ranking them based on a cost-function. Importantly, collision-avoidance and kinematic constraints were ignored during the trajectory generation process, leading to poor fulfilment of the given driving tasks. As a potential solution, we presented a trajectory optimizer that can generate a batch of solutions in parallel while incorporating all the necessary constraints. We showed that potential competing baselines based on state-of-the-art optimal control solver ACADO need up to 6x more computation time budget to produce comparable solution as ours. 

Our meta-costs can capture some of the rule-set described in \cite{rule_set_knoll}. For example, the velocity residual (\ref{cruise_cost}) can be easily modified to induce behaviors where ego-vehicle overtakes only when it leads to some minimum gain in the forward velocity. Similarly, "keep-right" rule from \cite{rule_set_knoll} is already incorporated in the meta-cost (\ref{hsrl_cost}). We present a more detailed analysis in our supplementary material \cite{github_repo_driving}.

\noindent \textbf{Limitations and Possible Workarounds:} Our batch optimization and MPC is capable of handling curved roads and residential driving scenarios (see accompanying video). However, it might struggle in highly cluttered and unstructured environments like parking lot that might require a large heading change. A possible workaround is to model the ego-vehicle geometry as a combination of circles, thus explicitly bringing the heading angle in the collision avoidance model \cite{aks_icra20}. Our preliminary results have shown that multi-circle approximation preserves the core batch structure at least for the holonomic robots \cite{aks_gpu_batch_2022}. 

Our batch optimizer structure, specifically the update rules (\ref{lin_solve_xy_2}),(\ref{lin_solve_psi_2}) is expected to be preserved for the more complex bi-cycle model of the ego-vehicle and we present a sketch of that derivation in the supplementary material \cite{github_repo_driving}. However, extension to include dynamics and tire forces will require major overhaul of the matrix algebra. 


\noindent \textbf{Future Work:} Our batch optimizer can be an attractive policy class for reinforcement learning algorithms. It can ensure safe exploration and thus would be beneficial for deep-Q learning based approaches.
Our batch MPC can also be used to generate supervision data for imitation learning algorithms like \cite{chauffeurnet}. Our future efforts are geared towards these directions.


\bibliography{main_ral}
\bibliographystyle{IEEEtran}

\end{document}